# Harnessing Artificial Intelligence to Infer Novel Spatial Biomarkers for the Diagnosis of Eosinophilic Esophagitis

◉ **Ariel Larey**[1,3,†]
Technion IIT

◉ **Eliel Aknin**[2,3,†]
Technion IIT

◉ **Nati Daniel**[3,†]
Technion IIT

◉ **Garrett A. Osswald**[4]
CCHMC

◉ **Julie M. Caldwell**[4]
CCHMC

◉ **Mark Rochman**[4]
CCHMC

◉ **Tanya Wasserman**[3]
Technion IIT

◉ **Margaret H. Collins**[5]
CCHMC

◉ **Nicoleta C. Arva**[6]
Northwestern University

◉ **Guang-Yu Yang**[7]
Northwestern University

◉ **Marc E. Rothenberg**[4]
CCHMC

◉ **Yonatan Savir**[3,*]
Technion IIT

## Abstract

Eosinophilic esophagitis (EoE) is a chronic allergic inflammatory condition of the esophagus associated with elevated esophageal eosinophils. Second only to gastroesophageal reflux disease, EoE is one of the leading causes of chronic refractory dysphagia in adults and children. EoE diagnosis requires enumerating the density of esophageal eosinophils in esophageal biopsies, a somewhat subjective task that is time-consuming, thus reducing the ability to process the complex tissue structure. Previous artificial intelligence (AI) approaches that aimed to improve histology-based diagnosis focused on recapitulating identification and quantification of the area of maximal eosinophil density. However, this metric does not account for the distribution of eosinophils or other histological features, over the whole slide image. Here, we developed an artificial intelligence platform that infers local and spatial biomarkers based on semantic segmentation of intact eosinophils and basal zone distributions. Besides the maximal density of eosinophils (referred to as Peak Eosinophil Count [PEC]) and a maximal basal zone fraction, we identify two additional metrics that reflect the distribution of eosinophils and basal zone fractions. This approach enables a decision support system that predicts EoE activity and classifies the histological severity of EoE patients. We utilized a cohort that includes 1066 biopsy slides from 400 subjects to validate the system's performance and achieved a histological severity classification accuracy of 86.70%, sensitivity of 84.50%, and specificity of 90.09%. Our approach highlights the importance of systematically analyzing the distribution of biopsy features over the entire slide and paves the way towards a personalized decision support system that will assist not only in counting cells but can also potentially improve diagnosis and provide treatment prediction.

---

[†]These authors have contributed equally to this work and share first authorship.
[*]Corresponding author, e-mail: yoni.savir@technion.ac.il
[1]Department of Computer Science, Technion IIT, Haifa 3200003, Israel. [2]Department of Industrial Engineering, Technion IIT, Haifa 3200003, Israel. [3]Department of Physiology, Biophysics and System Biology, Faculty of Medicine, Technion IIT, Haifa, Israel. [4]Division of Allergy and Immunology, Cincinnati Children's Hospital Medical Center, University of Cincinnati College of Medicine, Cincinnati, OH, USA. [5]Division of Pathology, Cincinnati Children's Hospital Medical Center, University of Cincinnati College of Medicine, Cincinnati, OH, USA. [6]Department of Pathology and Laboratory Medicine, Ann and Robert H. Lurie Children's Hospital of Chicago, Feinberg School of Medicine, Northwestern University, Chicago, IL, USA. [7]Department of Pathology, Northwestern University Feinberg School of Medicine, Chicago, IL, USA.

# 1 Introduction

Eosinophilic esophagitis (EoE) is a chronic immune system disease associated with esophageal tissue inflammation and injury characterized by a large number of eosinophils, which are found in the lining of the esophagus, called the esophageal mucosa [1]. EoE is allergy-driven and mainly caused by a reaction to food [2]. The damaged esophageal tissue leads to symptoms, such as pain and trouble swallowing [3]. In particular, EoE is becoming a more common cause of dysphagia in adults and vomiting, failure to thrive, and abdominal pain in children [3]. EoE can be treated by dietary restriction or topical steroids, and in more severe conditions, an endoscopic dilation intervention, specifically stricture dilation, is used.

Currently, the diagnosis of EoE relies on performing an upper endoscopy and obtaining esophageal mucosal biopsies. The hematoxylin and eosin (H&E) stained slides [4] are examined by pathologists. The physicians typically manually examine the slide using a microscope, identify the area of the tissue with the greatest eosinophil density, and count the number of intact eosinophils in that high-power field (HPF), i.e., the peak eosinophil count (PEC). The gold standard, histologic criterion, to date, is to define patients with EoE as having active disease if their $PEC \geq 15$ [5].

Yet, the PEC score captures only the maximal eosinophil count and not other properties such as the distribution of the eosinophils within the tissue, and it does not account for other cellular features that are captured by the EoE histology scoring system (EoEHSS) [6]. This method includes eight features that are relevant to EoE and accounts not only for the maximal severity of these features, but also for their distribution. This includes, for example, quantifying the percentage of HPFs within the slide that exceed the threshold of $\geq 15$ eosinophils. However, estimating such a metric visually poses a significant challenge. Another example of the importance of accounting for features in addition to the maximal eosinophil count is the development of a histological severity score that was used to diagnose remission (EoEHRS) [7]. In this case, both $PEC < 15/HPF$ and total grade and stage scores from all EoEHSS features $\leq 3$ are required to define remission.

Whereas processing the features of the entire whole slide improves diagnostic metrics, current manual approaches limit it. Counting PEC and scoring EoE histology is time-consuming, requires trained personnel, and can lead to variability between pathologists upon EoE biopsy diagnosis [8, 5, 9]. Hence, in recent years, considerable effort has been dedicated to build a robust and trustworthy process of inferring pathological biomarkers in health and disease. This includes harnessing machine learning in general and deep learning specifically [10, 11, 12, 13, 14, 15, 16, 17, 18, 19]. We have recently applied a dual approach towards diagnosing EoE: the first one is assigning a global label for the pathology images that is based on the patient condition [20]. The second one is based on segmenting and counting inflammatory cells, such as Intact eosinophils and Not-Intact eosinophils for EoE biopsy diagnosis using a deep convolutional neural network (DCNN) [21] .

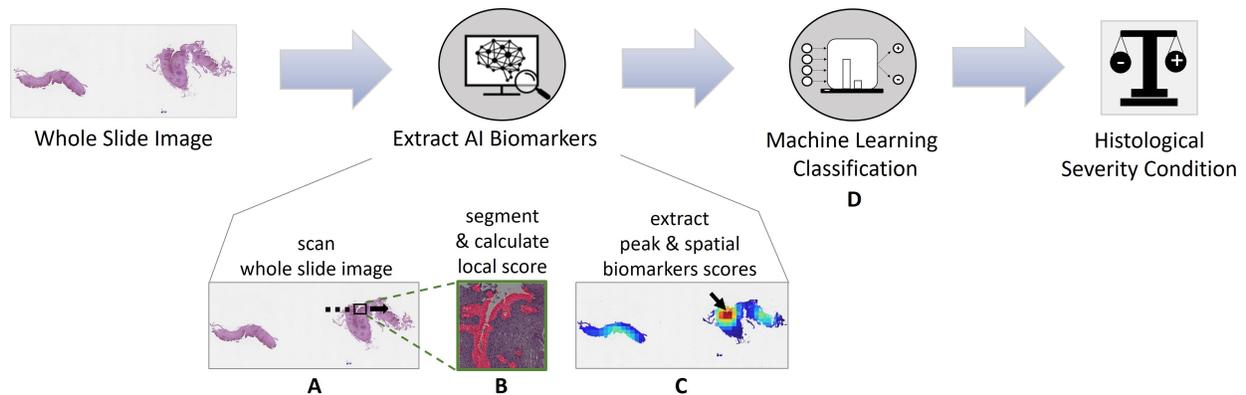

Figure 1: Artificial intelligence pipeline for diagnosing whole slide images (WSIs) and predicting disease activity of patients with eosinophilic esophagitis (EoE). (A) First, we analyze the WSI with a high-power-field (HPF)-sized kernel. (B) For each HPF, we segment intact eosinophils (Eos-intact) and basal zone (BZ) areas to obtain a local score for both features. (C) Once we have the analyzed entire WSI, we extract four biomarker scores that depend on the spatial distributions of eosinophils and basal zone. (D) We use these four biomarkers to predict the histological severity of the patients' conditions.

Here, we developed an artificial intelligence (AI) approach using machine learning for extracting novel biomarkers and used it to predict the histological severity condition (Figure 1). The pipeline has a state-of-the-art segmentation performance with a mean intersection over union metric (mIoU) score of 83.85% based on basal zone (BZ) and intact eosinophils (Eos-Intact) features. We show that derived biomarkers significantly correlate with manually obtained HSS scores. Using a cohort of 1066 biopsy slides from 400 patients, we demonstrate that AI biomarkers estimate histological severity achieving an accuracy of 86.70%, sensitivity of 84.50%, and specificity of 90.09%.

## 2 MATERIALS AND METHODS

### 2.1 Dataset and clinical scores

The dataset is part of the Consortium of Eosinophilic Gastrointestinal Disease Researchers (CEGIR) [22], a national collaborative network in the U.S. of 16 academic centers caring for adults and children with eosinophilic gastrointestinal disorders. The institutional review boards approved this study of the participating institutions via a central institutional review board at Cincinnati Children's Hospital Medical Center (CCHMC IRB protocol 2015-3613). Participants provided written informed consent. The dataset contains subjects with a history of EoE undergoing endoscopy (EGD) for standard-of-care purposes (n = 419). Distal, mid, or proximal esophageal biopsies (1-3 per anatomical site) per patient were placed in 10% formalin; the tissue was then processed and embedded in paraffin. Sections (4µm) were mounted on glass slides and subjected to hematoxylin and eosin (H&E) staining. Slides were scanned on the Aperio scanner at $400X$ magnification and were saved in SVS format. Each slide of esophageal tissue was analyzed by an anatomic pathologist who is a member of the CEGIR central pathology core. In addition to determining peak eosinophil count per 400X HPF (PEC), the pathologist subjected each slide to eosinophilic esophagitis histological scoring system (EoE HSS) analysis to assess the severity (grade) and extent (stage) of a set of histological abnormalities using a 4 point scale (0 normal; 3 maximum change) [6]. These features included eosinophilic inflammation (EI), basal zone hyperplasia (BZH), dilated intercellular spaces (DIS), eosinophilic abscess (EA), eosinophil surface layering (SL), surface epithelial alteration (SEA), dyskeratotic epithelial cells (DEC), and lamina propria fibrosis (LPF) [6]. The BZH grade score is determined by the amount of total epithelial thickness occupied by the basal zone, where 0 indicates that BZH is not present, 1 indicates that basal zone occupies >15% but <33% of the total epithelial thickness, 2 indicates that the basal zone occupies 33-66% of the total epithelial thickness, and 3 indicates that the basal zone occupies >66% of the total epithelial thickness. The BZH stage score indicates the amount of biopsy that indicated any degree of BZH, where 0 indicates that BZH is not present, 1 indicates that <33% of the epithelium exhibits any BZH with grade >0, 2 indicates that 33-66% of the epithelium exhibits any BZH with grade > 0, and 3 indicates that >66% of the epithelium exhibits any BZH with grade > 0 [6].

### 2.2 Semantic labeling

To train and validate the models, we labeled 23 patients' whole slide images (WSIs). The dataset consists of large WSIs with median length and width of 150,000 and 56,000 pixels, respectively. We cropped each WSI into small patches with a size of 1200 X 1200 pixels. Patches with a small amount of tissue, less than 15% of the patch area, were filtered. A total of n = 10,170 patches was used for semantic labeling. Those patches were analyzed and annotated by an expert using VIA [23] and then were verified by three different experts. For each patch, the intact eosinophils' centers and the basal zone area were marked. The result was two semantic masks. In the first, the pixels in the area of a circle with a radius of 25 pixels around the intact eosinophils center were labeled as Eos-Intact [21]. In the second, pixels within the marked basal zone polygons were labeled as BZ. That is, each pixel was classified either as a BZ type, Eos-Intact type, both of them, or as none. In total, about 570 million pixels were labeled as BZ, and about 78.47 million pixels were labeled as Eos-Intact. 8.6% of the images contained BZ, where their area was, on average, 45.45% of the image size. Eos-Intact were found in 22.8% of the images, with an average area fraction of 2.35%.

### 2.3 Semantic segmentation

We trained two models, one using the Eos-Intact masks and one using the BZ masks. For both models, the annotated patches were divided into two groups; 80% of the data were dedicated to training the segmentation model, and the rest, 20%, for testing the model. The segmentation model was based on the UNet++ architecture [24]. It was developed in the PyTorch framework [25] and was trained on a single NVIDIA GeForce RTX 2080 Ti GPU. During the training phase, the 1200 X 1200-pixel patches were divided into 448 X 448-pixel sub-patches with an overlap of 72 pixels between them. Different sub-patch sizes were tested, and this size was optimal in terms of precision and recall (see segmentation metrics section of the systems). In addition, multiple hyperparameters were tested. The optimal parameters were batch size of 5, "Cosine Annealing" learning rate scheduler, and a 0.5 softmax threshold. The optimization loss function contains two terms, the Dice and Binary cross-entropy (BCE), where each term is weighted. After exploring different weights, we applied the weights 1 and 0.5 to the Dice and BCE, respectively. For inference, the test image was cropped into 448 X 448-pixel sub-patches as described above. To reduce segmentation noise, contiguous regions labeled as Eos-Intact or BZ that were smaller than an area of 1800 pixels, in the case of Eos-Intact, or area of 2007 (1% out of the sub-patch size), in the case of BZ, were re-labeled as none.

## 2.4 Semantic metrics

To estimate the segmentation performances, we used the following metrics,

$$mIoU = \frac{1}{I \cdot C} \sum_i \sum_c \frac{TP_{i,c}}{TP_{i,c} + FP_{i,c} + FN_{i,c}} \quad (1)$$

$$mPrecision = \frac{1}{I \cdot C} \sum_i \sum_c \frac{TP_{i,c}}{TP_{i,c} + FP_{i,c}} \quad (2)$$

$$mRecall = \frac{1}{I \cdot C} \sum_i \sum_c \frac{TP_{i,c}}{TP_{i,c} + FN_{i,c}} \quad (3)$$

$$mSpecification = \frac{1}{I \cdot C} \sum_i \sum_c \frac{TN_{i,c}}{TN_{i,c} + FP_{i,c}} \quad (4)$$

where the $c$ index iterates over the different classes in the image, and the $i$ index iterates over the different images in the dataset. $C$ is the total number of classes, and $I$ is the total number of images. $TP$, $TN$, $FP$, and $FN$ are classification elements that denote true positive, true negative, false positive, and false negative of the areas of each image, respectively.

| Metric | Eos-Intact | BZ | Overall |
| --- | --- | --- | --- |
| mIoU (Equation 1) | 0.93 | 0.75 | 0.84 |
| mPrecision (Equation 2) | 0.95 | 0.8 | 0.88 |
| mRecall (Equation 3) | 0.97 | 0.94 | 0.95 |
| mSpecificity (Equation 4) | 0.998 | 0.82 | 0.91 |

Table 1: Four segmentation metrics measured at the pixel level. IoU denotes the Intersection Over Union between the Ground Truth and the prediction. Recall denotes the fraction of the True-Positive pixels among the total Ground Truth pixels in the image, whereas Precision denotes the fraction between the True-Positive pixels and the prediction pixels. The fraction between the True-Negative pixels and the total negative pixels in the image is coined Specificity. mIoU, mRecall, mPrecision and mSpecificity are obtained by averaging IoU, Recall, Precision, and Specificity, respectively, over the validation set. The metrics are presented for the Eos-Intact and BZ classes separately in addition to their average per image as the overall score. The compared patches size is the network's input size - 448 X 448 pixels.

## 2.5 Calculating WSI AI scores

To evaluate the eosinophil and basal zone distribution within each WSI, we use an iterative process to scan over the entire slide. At each step, an image the size of a HFP is processed. The area of an HPF corresponds to a size of 2144 X 2144 pixels (548 µm X 548 µm). The stride step between constitutive HPFs is 500 pixels. Each HPF is divided into 25 sub-patches (448 X 448 pixels - corresponding to the network input size) with an overlap of 24 pixels. Each sub-patch is segmented and the HPF segmentation mask is assembled from them. The pixels' identity in the areas overlapping between sub-patches is determined by using OR function. After segmentation, each HPF is assigned two local scores: the number of intact eosinophils [21] and the BZ area rate, which is the ratio of the number of BZ pixels in the HPF mask, to the HPF size. After scanning the entire WSI, we produce score maps for both features - an Intact-Eosinophils map and a BZ map, where every pixel in these maps represents the score of the matching HPF. Based on the score maps, we can produce four WSI scores (Figure 1C):

- Peak Eosinophil Count (PEC) - The number of eosinophils in the HPF with the densest area of eosinophils within the WSI. This score is used in the clinic to diagnose active EoE [5, 21]. A patient with a PEC greater than or equal to 15 is considered to have active EoE. The EI grade score is a proxy for this measure.
- Spatial Eosinophil Count (SEC) - The ratio of the number of HPFs with an Intact-Eosinophil count that is greater than or equal to 15 to the total number of HPFs in the feature map. The EI stage score is a proxy for this measure.

- Peak Basal Zone (PBZ) - The maximum HPF BZ area rate. This score is the maximal density of basal cells per HPF in the WSI. The BZH grade score is a proxy for this measure.

- Spatial Basal Zone (SBZ) - The ratio of the number of HPFs with local BZ score that is greater than or equal to 15% to the number of tissue HPFs in the feature map. The BZH stage score is a proxy to this measure.

## 2.6 Classifying whole slide image

### 2.6.1 Features-based classification

We previously presented a pipeline for classifying WSIs using only the predicted PEC directly [21]. In this paper, we leverage the spatial information, for both eosinophils and basal cells that was revealed by segmenting the entire WSI. We used this information to devise four WSI scores and to predict the histological severity condition of the patient (Figure 1D). We explored different machine learning models - support vector machine (SVM), and linear discriminant analysis (LDA). In addition, various architectures of multi-layer perceptron (MLP) were examined, particularly, all combinations of layers in the size of 10, 20, 50, 100 tiled up to four hidden layers. We used these types of classifiers because of their better capability to handle tabular data (in contrast to convolutional-neural-networks, for example, that support sequential data). The cohort contains 1066 WSIs that were not used for the segmentation training. Classifier training was done using 80% of the data, whereas the rest were used for validation. For each model, we repeated the training procedure 20 times with different random seeds for splitting the data, and reported the median results.

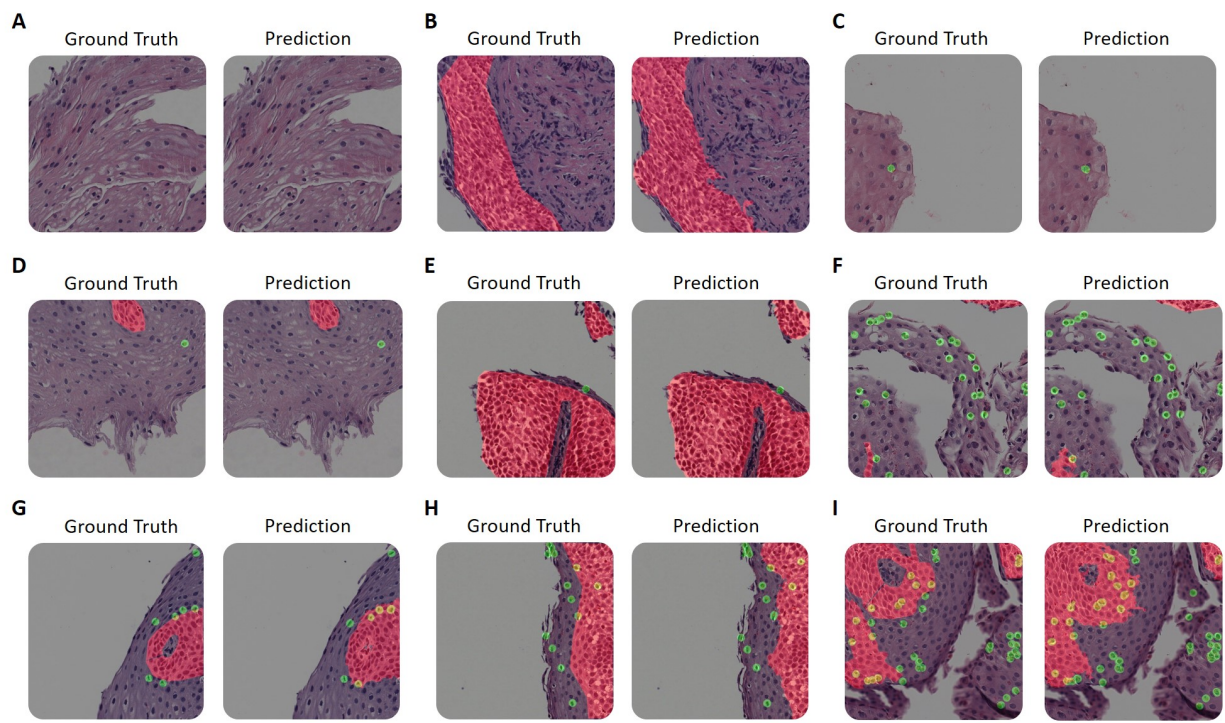

Figure 2: Examples of our platform semantic segmentation. (A-I) The size of each image is 1200 X 1200 pixels. Each panel's left-hand side is colored according to the ground truth as annotated by trained experts. The right-hand side is colored with its corresponding network prediction mask. Basal zone (BZ) pixels are colored with red, intact eosinophils (Eos-Intact) pixels are colored with green, and pixels associated with both (that is, eosinophils within a BZ area) are colored with yellow. (A-C) The upper row shows examples with only one label or none. (D) An example of an image that contains both a small number of basal cells and intact eosinophils. (E) An example of an image with a large basal zone and a small number of intact eosinophils. (F) An example that contains a small area of basal zone and a large number of intact eosinophils. (G-I) The bottom row displays examples with large basal zones and also a large number of intact eosinophils.

### 2.6.2 Multi-classification

To improve the histological severity classification performance, different classifiers were used for regions having different eosinophil density. We define two regions of PEC scores,

$$classifier = \begin{cases} C_{in} & (PEC \geq 15 - \Delta) \text{ and } (PEC \leq 15 + \Delta) \\ C_{out} & (PEC < 15 - \Delta) \text{ or } (PEC > 15 + \Delta) \end{cases} \quad (5)$$

where $C_{in}$ and $C_{out}$ denote the classifier inside the window and outside of the window, respectively. The hyperparameter $\Delta$ defines the window size. The training procedure is as described above. To avoid bias, the contribution of each region to the 80%-20% split is proportional to the region size, ensuring that each region contributes points to the training and validation. We examined $\Delta$ values in the range of $[1, 12]$.

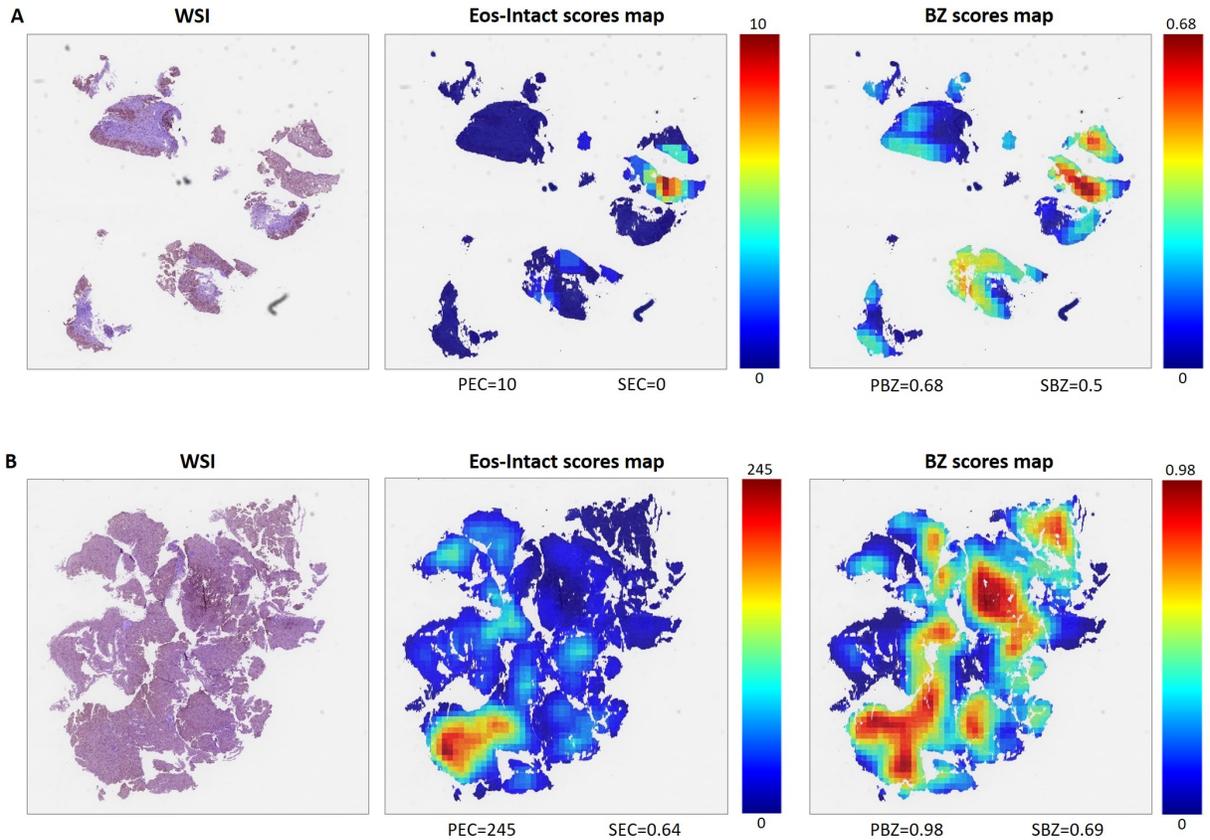

Figure 3: Examples of two different WSIs (left) and their corresponding scores maps with scale for each score defined (middle, right). Each pixel in these maps represents one HPF, and the color of the pixel indicates the respective score. From the Eos-Intact scores map (middle), we extracted peak eosinophil count (PEC) and spatial eosinophil count (SEC). From the basal zone (BZ) score map (right), we computed peak basal zone (PBZ) and spatial basal zone (SBZ) scores. (A) Example of a WSI of a biopsy obtained from an EoE patient with inactive disease (PEC=10). (B) Example of a biopsy obtained from a patient with active EoE (PEC = 245).

## 3 RESULTS

### 3.1 Local segmentation results

Figure 2 illustrates a few examples of our platform semantic segmentation compared with ground truth labeling by a trained researcher. Table 1 summarizes the segmentation metrics over the whole validation-set, 1, 2, 3, and 4.

### 3.2 WSI features scores

One of the main advantages of the described approach is that it allows scoring that is based not only on a limited number of regions probed by the pathologist but on the entire whole slide image (Fig. 3). To process the entire whole slide image, we used dynamics convolution to scan the slide using windows with a HPF size with a stride of about 1/4 of the HPF size (Subsection 2.5). We computed the score maps for 1066 WSIs from 400 patients that were not part of the semantic segmentation training and validation sets. The pipeline produces two feature-score maps for each WSI, one for the Eos-Intact score map and the second for the BZ score map. Figure 3 shows examples of two features score maps computed from two different WSIs. We computed four scores based on the semantic segmentation of the WSI; this included two local ones (peak eosinophil counts [PEC] and peak basal zone [PBZ]), and two global ones (spatial eosinophil counts [SEC] and spatial basal zone [SBZ]) (Subsection 2.5). We compared the different WSI scores with the relevant HSS score estimated by the pathologists. We compared PBZ, SBZ, PEC, and SEC with HSS BZH grade, HSS BZH stage, HSS EI grade and HSS EI stage, respectively (Subsection 2.5). Our scores showed a significant correlation with the human estimated metrics (Fig. 4A-D). We then analyzed the relationship between the two types of biomarkers: the number of eosinophils and the area of the basal zone. It was suggested that these features have some correlation between them [6]. A standard condition for the classification of a patient as having active EoE is having a PEC that is greater than or equal to 15. We show that the PBZ distribution of non-active patients has significantly lower values than the PBZ score distribution of the active patients (Fig. 4E). A similar trend is observed when analyzing the SBZ distribution (Fig. 4F). Yet, there are still patients with high PEC scores and low PBZ / SBZ scores, and vice-versa. This raises the question of whether a combination of basal zone-based metrics can better predict the patient clinical status and treatment outcome.

| INPUT - WSI AI features scores | | | | OUTPUT - Classification Models results | | |
|---|---|---|---|---|---|---|
| | | | | SVM | LDA | MLP |
| PEC | SEC | PBZ | SBZ | med / std | med / std | med / std |
| + | + | | | 0.8364 / 0.0247 | 0.75 / 0.0787 | 0.8388 / 0.027 |
| | | + | + | 0.7991 / 0.0236 | 0.8061 / 0.0227 | 0.806 / 0.0227 |
| + | + | + | + | 0.8341 / 0.0233 | 0.8155 / 0.0208 | 0.8505 / 0.0285 |

PEC, Peak Eosinophil Count; SEC, Spatial Eosinophil Count; PBZ, Peak Basal Zone; SBZ, Spatial Basal Zone.
SVM, Support Vector Machine; LDA, Linear Discriminant Analysis; MLP, Multi-Layer Perceptron

Table 2: Classification results of multiple models (SVM, LDA and MLP) with different combinations of input features (PEC, SEC, PBZ, and SBZ). Each model was trained and validated 20 times with different train-validation random splits, the median (med) results are reported with the standard deviation (std).

### 3.3 Histological severity classification

The naive approach for diagnosing patients' histological severity condition uses only PEC information. In this approach, if the patient's PEC is greater than or equal to 15, the patient is considered to have active EoE. Similar criteria are also applied to determine whether a patient who underwent treatment responded and is in remission. Recent studies suggested using basal zone histological information improves the estimation of the disease's histological severity. For example, it was suggested that patients with low PEC values, i.e., greater than 0 but less than 15, but with basal zone hyperplasia would not be considered as patients in remission [7]. To test the performance of our pipeline in integrating all four WSI scores, we used as the ground truth (GT) a standard clinical histological severity metric that defines a histologically severe patient as one who is not in histologic remission, i.e., that has a PEC of greater than or equal to 15 or an HSS total score of more than 3 [7]. This metric is stringent when examining whether a patient is in remission or not compared to taking into account only the PEC score.

First, as a baseline classifier, we calculated the accuracy of the histological severity classification when it was based only on the PEC score. The best accuracy (83.3%) was obtained when the threshold criteria was PEC = 6. We recently showed that when taking only PEC as a metric for classification of the patient state (i.e., active EoE vs. non-active EoE), the AI-based PEC score provides a classification accuracy of 94.75%. Moreover, the optimal PEC threshold that provided the best accuracy in that case was 15 [21], the same as the gold standard threshold [5]. Thus, the current results suggest that to compensate for the cases in which low PEC are still considered histologically severe, the system converges to more tight PEC criteria for histological severity classification.

Next, we trained a classifier that takes into account all four metrics we calculated from the WSI score maps (i.e., PEC, SEC, PBZ, SBZ). We used several training approaches: support vector machine (SVM), linear discriminant analysis (LDA), and multi-layer perceptron (MLP). The best results were obtained using MLP with three hidden layers where each layer has 20, 50 and 100 neurons, respectively. Integrating all the metrics yields an improvement in accuracy to 85.05%. Moreover, the false alarm rate decreased by about 20% compared to the baseline classifier, whereas the miss rate decreased by about 5% (Fig. 5).

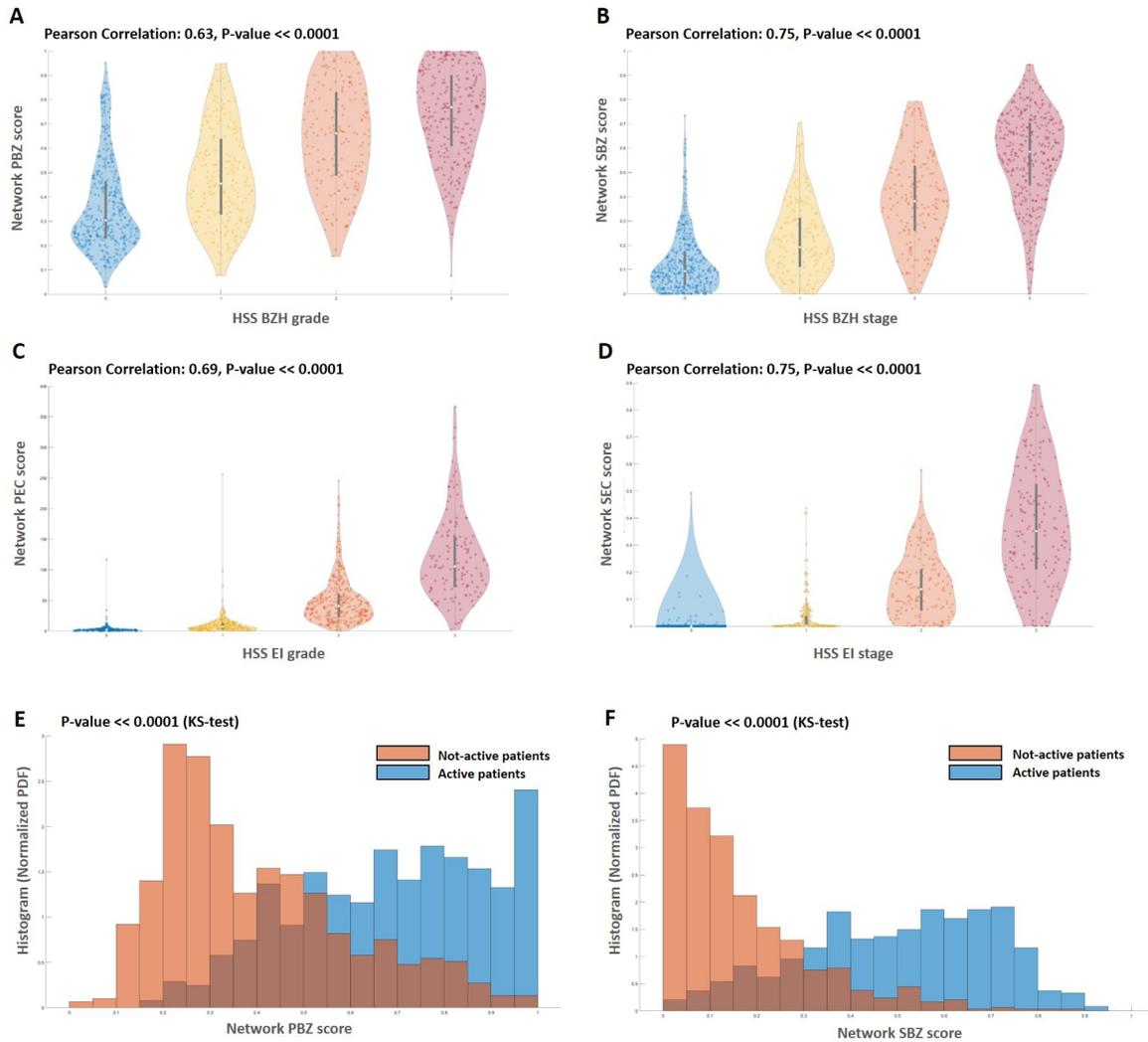

Figure 4: Correlations among the different score types. (A-D) Comparing the computed scores with the HSS scores. The HSS scoring method for BZH grade, BZH stage, EI grade, and EI stage, each score is an integer between zero and three. Each panel depicts a violin plot that shows the distribution of the computed WSI scores (vertical axis) for each HSS score that is the appropriate proxy (horizontal axis). The white circle indicates the median value, and the black bar indicates the standard deviation. There is a significant correlation between the computed scores and their HSS counterparts. (E-F) Histograms of basal zone related metrics PBZ (E) and SBZ (F) for active ($PEC \geq 15$) and non-active patients ($PEC < 15$). Both the PBZ and SBZ distribution scores of non-active patients have significantly lower values than the PBZ and SBZ distribution scores of the active patients. (Kolmogorov–Smirnov-test, P-value $<<$ 0.0001)

A possible factor that may impede the prediction performances is the fact that our data contain patients with a large range of eosinophil counts. To further improve the prediction, we took a multi-classification approach at which patients with a PEC level that is near the decision threshold are classified separately from patients that have a PEC level that is far from it. The best results were achieved when patients with PEC values within the range [6 24] were analyzed separately (Subsection 2.6.2). This approach led to an accuracy of 86.70% and a significant reduction in the false-alarm rate to 9.91% (Fig. 5). In this case, the best results were given by an MLP with three hidden layers in the size of 100, 20, and 100, respectively, for both classifiers.

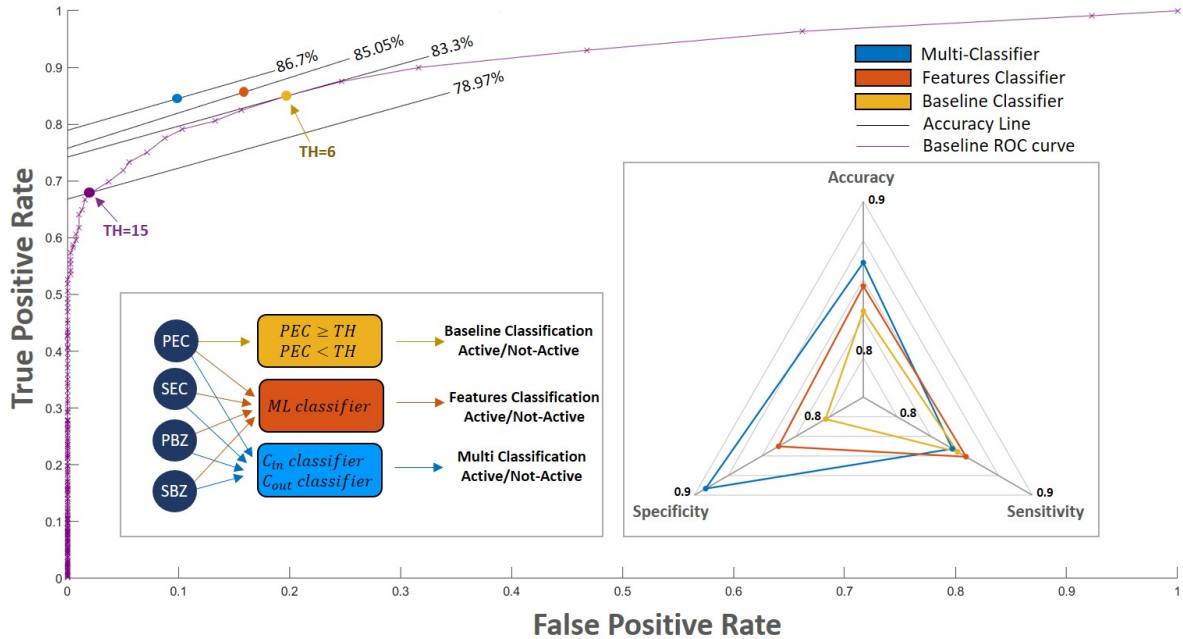

Figure 5: Classification performance of the different models. We examined a few different classification approaches (left inset): a baseline in which the classification is according only to the PEC score (yellow, ROC curve in purple), a trained classifier that accounts for all the four WSI scores (orange), and a multi-classification approach that separates between patients close to the decision threshold from those that are far from it (blue). Spider plot (right inset) depicts the performance of the different models. Accounting for all the AI WSI scores significantly improves the classification performance models. The multi-classification approach that separates patients near the decision threshold improves the performance even more.

To gain insight into the role of each of our four WSI scores, we explored the effect of training a classifier with a limited subset of them (Table 2). In all configurations, the best accuracy was obtained by the MLP model. As expected, the highest classification score was achieved when we used all four WSI AI features scores. Yet, accounting only of Eos-intact scores (PEC and SEC) provides better accuracy than using only BZ scores (PBZ and SBZ).

## 4 DISCUSSION

Biopsy-based diagnosis often requires the identification of features that are on the single-cell scale. In the case of EoE, the diagnosis procedure involves counting eosinophils and estimating their density. As a typical whole slide image contains at least tens of high-power fields, gold standard scores usually do not account for the entire features distribution. In the case of EoE the gold standard score takes into account only the maximal density of cells. One of the promises of digital pathology, besides automating manual tasks, is the ability to process the entire WSI and infer novel biomarkers that capture the spatial distribution of the relevant features.

In this work, we introduce an artificial intelligence system that infers novel local and spatial biomarkers based on semantic segmentation of intact eosinophils and basal zone. This approach enables a decision support system that takes into account information from the entire WSI and classifies EoE patients' histological severity. In previous works [21], we introduced a platform that infers the maximal eosinophil density and, based on that, predicts whether a patient has active disease or not with an accuracy of 94.75%. Here, we develop a platform that not only recapitulates the metric used by the pathologists but also provides novel biomarkers. Besides a metric that captures the maximal density of eosinophils (PEC) and the maximal basal zone fraction (PBZ), we suggest two additional metrics that reflect the distribution of eosinophils and basal zone fractions (SEC and SBZ respectively).

To test the platform, we utilized a cohort that includes 1066 biopsy slides from 400 subjects. Whereas the decision of whether EoE is active or not depends on a gold standard cutoff of 15 eosinophils per high power field, the histological severity score (mainly used to estimate whether a patient was in histologic remission after a treatment) also accounts for the basal zone properties. Indeed, using only the PEC of greater than or equal to 15 as a threshold to predict histological severity yields an accuracy of only 78.97%. The PEC cutoff that provides the best accuracy for histological severity, which was 83.3%, is 6 eosinophils/HPF. This reflects the fact that adding the basal zone criteria results in a stronger criteria for the PEC.

To improve the performance, we used a few machine learning approaches that take our metrics as an input. We show that taking the eosinophils metrics alone yields an accuracy of 83.4% whereas taking the basal zone metrics alone gives an accuracy of 80.6%. Putting all the metrics together gives an accuracy of 85.05%. That is, using all the metrics together gives better performances than each of the metrics alone and also better than a naïve approach of changing the PEC cutoff. Finally, we also constructed a multi classifier approach that is based on the fact that patients around the $PEC = 15$ cutoffs are more prone to errors. Altogether, our platform yields a classification accuracy of 86.70%, sensitivity of 84.50%, and specificity of 90.09%. Our approach highlights the importance of systematically analyzing the distribution of biopsy features over the entire slide image and putting together metrics based on them. Our platform paves the way towards a personalized decision support system that will assist in not only counting cells but also in providing treatment prediction.

## Author Contributions

YS and MER conceived and designed the research. YS and AL designed the pipeline. YS, AL, EA, and ND designed and coded the platform code. AL, EA, ND, and TW analyzed the data, and validated it. AL, EA, ND, and YS performed all the mathematical analyses. GO and JC contributed to the pipeline clinical aspects, annotated, and validated the segmentation data. GO, and JC organized and analyzed the data. MC, NA and GY annotated the CEGIR slides. MC supervised the data annotation. MC, MR, NA and GY, contributed to the pipeline clinical aspects. YS, AL, EA, ND, TW, and JC wrote the draft of the paper, which was reviewed, modified, and approved by all other authors.

## Funding

YS was supported by Israel Science Foundation #1619/20, Rappaport Foundation, and the Prince Center for Neurodegenerative Disorders of the Brain 3828931. M.E.R. was supported by NIH R01 AI045898-21, the CURED Foundation, and Dave and Denise Bunning Sunshine Foundation. CEGIR (U54 AI117804) is part of the Rare Disease Clinical Research Network (RDCRN), an initiative of the Office of Rare Diseases Research (ORDR), NCATS, and is funded through collaboration between NIAID, NIDDK, and NCATS. CEGIR is also supported by patient advocacy groups including American Partnership for Eosinophilic Disorders (APFED), Campaign Urging Research for Eosinophilic Diseases (CURED), and Eosinophilic Family Coalition (EFC). As a member of the RDCRN, CEGIR is also supported by its Data Management and Coordinating Center (DMCC) (U2CTR002818).